\documentclass[letterpaper]{article} 
\usepackage[submission]{aaai23}  
\usepackage{times}  
\usepackage{helvet}  
\usepackage{courier}  
\usepackage[hyphens]{url}  
\usepackage{graphicx} 
\usepackage{natbib}  
\usepackage{caption} 
\usepackage{xcolor}
\usepackage{epsfig}
\usepackage{amsmath}
\usepackage{amssymb}
\usepackage{bm}
\usepackage{amsfonts} 
\usepackage{makecell}
\usepackage{multirow}
\usepackage{subcaption}
\usepackage{amsthm}
\usepackage{booktabs}
\usepackage{algorithm}
\usepackage{algorithmic}
\usepackage{soul}
\usepackage{pifont}

\newcommand{\bmmc}[1]{\bm{\mathcal{#1}}}
 
\newcommand{\cmark}{\ding{51}}%
\newcommand{\xmark}{\ding{55}}%

\usepackage{newfloat}
\usepackage{listings}
\floatstyle{ruled}
\newfloat{listing}{tb}{lst}{}
\floatname{listing}{Listing}
\title{HALOC: Hardware-Aware Automatic Low-Rank Compression for Compact Neural Networks}
\author {
    Jinqi Xiao\textsuperscript{\rm 1}, 
    Chengming Zhang\textsuperscript{\rm 2}, 
    Yu Gong\textsuperscript{\rm 1}, 
    Miao Yin\textsuperscript{\rm 1}, \\
    Yang Sui\textsuperscript{\rm 1}, 
    Lizhi Xiang\textsuperscript{\rm 3}, 
    Dingwen Tao\textsuperscript{\rm 2,\rm 3},
    Bo Yuan\textsuperscript{\rm 1}
}
\affiliations {
    \textsuperscript{\rm 1} Rutgers University, \textsuperscript{\rm 2} Indiana University, \textsuperscript{\rm 3} Washington State University\\
    jinqi.xiao@rutgers.edu, czh5@iu.edu, yg430@soe.rutgers.edu, miao.yin@rutgers.edu,\\ yang.sui@rutgers.edu, lizhi.xiang@wsu.edu, ditao@iu.edu, bo.yuan@soe.rutgers.edu
}

\begin{document}

\maketitle

\begin{abstract}
Low-rank compression is an important model compression strategy for obtaining compact neural network models. In general, because the rank values directly determine the model complexity and model accuracy, proper selection of layer-wise rank is very critical and desired. To date, though many low-rank compression approaches, either selecting the ranks in a manual or automatic way, have been proposed, they suffer from costly manual trials or unsatisfied compression performance. In addition, all of the existing works are not designed in a hardware-aware way, limiting the practical performance of the compressed models on real-world hardware platforms.  

To address these challenges, in this paper we propose HALOC, a hardware-aware automatic low-rank compression framework. By interpreting automatic rank selection from an architecture search perspective, we develop an end-to-end solution to determine the suitable layer-wise ranks in a differentiable and hardware-aware way. We further propose design principles and mitigation strategy to efficiently explore the rank space and reduce the potential interference problem.

Experimental results on different datasets and hardware platforms demonstrate the effectiveness of our proposed approach.  On CIFAR-10 dataset, HALOC enables 0.07\% and 0.38\% accuracy increase over the uncompressed ResNet-20 and VGG-16 models with 72.20\% and 86.44\% fewer FLOPs, respectively. On ImageNet dataset, HALOC achieves 0.9\% higher top-1 accuracy than the original ResNet-18 model with 66.16\% fewer FLOPs. HALOC also shows 0.66\% higher top-1 accuracy increase than the state-of-the-art automatic low-rank compression solution with fewer computational and memory costs. In addition, HALOC demonstrates the practical speedups on different hardware platforms, verified by the measurement results on desktop GPU, embedded GPU and ASIC accelerator. 
\end{abstract}

\section{Introduction}

\textit{Model compression} is an important deep neural network (DNN) optimization strategy that aims to improve the execution efficiency of deep learning. Motivated by the observation that considerable redundancy exhibits at different levels (e.g., neuron and bit) of DNNs, various compression approaches, such as pruning \cite{han2015learning,dong2020rtmobile, 9499915} and quantization \cite{rastegari2016xnor, zhang2018lq}, have been widely studied and developed to reduce model redundancy with different granularity.

Among the existing compression methods, \textit{low-rank compression} is a unique solution that explores the low-rankness of DNN model at the structure level. By decomposing the original large-size weight matrices or tensors to a series of small cores, low-rank compression can bring significant storage and computational savings. To date, many decomposition-based compression approaches have been proposed. Based on their differences in the factorization schemes, they can be categorized to 2-D \textit{matrix decomposition} based \cite{klema1980singular} and high-order \textit{tensor decomposition} based \cite{tucker1966some, harshman1970foundations}.

\textbf{Low-Rank Compression in Practice.} From the perspective of real-world deployment, practical low-rank DNN compression should satisfy two requirements: \textit{\textbf{Automatic Rank Selection}} and \textit{\textbf{Hardware-awareness}}. \ul{First}, assume that using Tucker-2 decomposition \cite{kim2016compression} to compress an $N$-layer DNN with maximum rank value as $M$ per layer, there are total $M^{2N}$ possible rank settings. Consider in practice $N$ is typically tens to hundreds and $M$ can be up to tens, the overall rank selection space is extremely huge. Evidently, automatic rank selection is highly desired to reduce the expensive search efforts and the potential sub-optimality incurred by the heuristic manual setting. \ul{Second}, because the decomposed DNN models are essentially executed on the practical computing hardware, low-rank compression should be performed towards reducing the hardware-cost metrics, e.g., latency and energy, to obtain the actual benefits in the real-world scenarios. Also, consider the large diversity of current DNN computing hardware (e.g., desktop GPU, embedded GPU, application-specific integrated circuit (ASIC) accelerator, etc.) and different platforms may favor different network structures, the direct feedback from the target devices should be included in the low-rank compression process \cite{xiang2022tdc}.

\textbf{Limitations of Existing Works.} Measured from the above two practical requirements, the existing low-rank DNN compression approaches have several limitations. \ul{First}, most of the prior works determine the ranks via rounds of manual trials and adjustments, causing expensive engineering efforts. \ul{Second}, even for the state-of-the-art solutions with automatic rank selection, due to the insufficient exploration of rank space, their compression performance is still unsatisfied, sometimes even lower than the approaches with heuristic rank settings. \ul{Third}, all of the existing low-rank decomposition approaches, no matter determining the ranks manually or automatically, do not consider hardware-awareness in the design process, limiting their performance for deployment on practical hardware platforms.


\textbf{Technical Contributions and Preview.} To address these challenges and promote the widespread adoption of low-rank compression in practice, in this paper we propose HALOC, a hardware-aware automatic low-rank compression framework for compact DNN models. Based on the observation that rank selection process essentially determines the architecture of the low-rank models, we interpret automatic rank selection as the automatic search for the proper low-rank structure, enabling efficient automatic low-rank compression in an differentiable and hardware-aware way. Overall, the contributions of this paper is summarized as follows:
\begin{itemize}
    \item We identify the hidden connection between setting the layer-wise ranks and searching the network architecture. Based on this interesting discovery, we propose to develop a new low-rank compression strategy with neural architecture search (NAS)-inspired automatic rank selection scheme, which can sufficiently explore the rank space and strong compatibility for hardware-awareness.
    
    \item We propose two low-rank-specific design principles to realize efficient exploration in the rank space, leading to significant reduction in the search cost. We also analyze the potential methods for mitigating the interference problem in the search process, and identify the suitable solution to improve compression performance.
    
    \item We perform evaluation experiments of compressing various models on different datasets, and also measure the practical speedup across different hardware platforms. On CIFAR-10 dataset, HALOC achieves 0.07\% and 0.38\% accuracy increase over the original uncompressed ResNet-20 and VGG-16 models with 72.20\% and 86.44\% FLOPs reduction, respectively. On ImageNet dataset, HALOC brings 0.9\% top-1 accuracy increase over the original ResNet-18 model with 66.16\% FLOPs reduction. When compressing MoblieNetV2 on ImageNet dataset, HALOC shows 0.66\% top-1 accuracy increase over the state-of-the-art low-rank compression method with fewer FLOPs and memory cost. The measurement results on various hardware platforms (desktop GPU, embedded GPU and ASIC accelerator) demonstrate the practical speedups brought by our proposed hardware-aware solution.
    
\end{itemize}

\section{Related Work}

\textbf{Low-Rank DNN Compression.} Exploring structure-level low-rankness has been well studied for neural network compression. In general, a compact DNN can be obtained via performing low-rank matrix or tensor decomposition on the original large model. For matrix factorization-based methods \cite{denton2014exploiting,wen2017coordinating,8942082,idelbayev2020low,liebenwein2021compressing}, all the weight tensors, including the 4-D type for the convolutional (CONV) layers, are first flatten to 2-D format and then decomposed to two small matrices; while tensor decomposition-based approaches, including Tucker \cite{kim2016compression,gusak2019automated,yin2020compressing, yin2021towards}, CP \cite{lebedev2014speeding,astrid2017cp}, tensor train \cite{novikov2015tensorizing,wang2019lossless, deng2019tie, Yin_2022_CVPR} and tensor ring \cite{pan2019compressing}, directly factorize the high-order tensor objectives to a series of tensor and matrix cores. No matter which specific decomposition is adopted, one key challenge is the efficient rank setting. Currently most of the existing works determine the layer-wise ranks via hand-crafted manual trials and attempts, a strategy that requires very expensive engineering efforts.

\textbf{Automatic Rank Selection.} Our work is most closely related to \cite{gusak2019automated,liebenwein2021compressing,li2022heuristic, yin2022batude}, the state-of-the-art automatic rank selection solutions. Specifically, \cite{gusak2019automated} proposes to utilize variational Bayesian matrix factorization to determine the ranks of the tensor decomposed DNNs in a multi-stage way. In \cite{liebenwein2021compressing}, Eckhart-Young theorem \cite{golub2013matrix} is used to determine the layer-wise ranks of the factorized weight matrices towards minimizing compression error. Besides, \cite{li2022heuristic} applies genetic algorithm to the rank search process of tensor ring decomposed DNNs. Compared with the manual trials, these automatic selection approaches indeed facilitate the rank determination procedure. However, their accuracy performance is still limited because of the insufficient exploration of rank space. Also, similar to the existing manual setting-based solutions, these automatic compression approaches are not designed in a hardware-aware way, limiting the performance of the compressed models on the practical hardware platforms.

\section{Preliminaries}


\textbf{Notation.} We denote a tensor by using boldface calligraphic script letter, e.g., $\bmmc{X}$. Matrices and vectors are represented in the format of boldface capital letters and lower-case capital letters, e.g. $\bm{X}$ and $\bm{x}$, respectively. In addition, non-boldface letters with indices $\mathcal{X}(i_1,\cdots,i_d)$, $X(i,j)$ and $x(i)$ represent the entries for $d$-dimensional tensor $\bm{\mathcal{X}}$, matrix $\bm{X}$ and vector $\bm{x}$, respectively.\\
\textbf{Tucker-2 Decomposed CONV Layer.} Without loss of generality, in this paper we study the automatic low-rank compression using Tucker-2 decomposition. In general, for a CONV layer with weight tensor $\mathcal{W} \in \mathbb{R}^{F \times C \times K_1 \times K_2}$, 
where $C$ is the number of input channels, $K_1$ and $K_2$ are the kernel size, and $F$ is the number of output channels, it can be factorized to a core tensor and two matrices along each mode of Tucker-2 decomposition as follows:
\begin{equation}
\label{eq:tucker2}
\begin{aligned}
&\mathcal{W}(f,c,i,j) =\sum_{{r_1}=1}^{r^{(1)}} \sum_{{r_2}=1}^{r^{(2)}} \mathcal{C}({r_1},{r_2},i,j) M_1({r_1},f) M_2({r_2},c),
\end{aligned}
\end{equation}
where $\bm{\mathcal{C}} \in \mathbb{R}^{r^{(1)}\times r^{(2)}\times K_1\times K_2}$, $\bm{M}_1 \in \mathbb{R}^{r^{(1)}\times F}$, $\bm{M}_2 \in \mathbb{R}^{r^{(2)}\times C}$, and $r^{(1)}$ and $r^{(2)}$ denote the tensor rank. Then given an input tensor $\bmmc{X} \in \mathbb{R}^{W\times H\times C}$ and an output tensor $\bmmc{Y} \in \mathbb{R}^{W'\times H'\times F}$, the Tucker-2-format convolution is performed as follows:
\begin{equation}
\begin{aligned}
\label{eq:step1}
\mathcal{K}_1(w,h,{r_2}) = \sum_{c=1}^{C} M_2({r_2},c)\mathcal{X}(w,h,c),
\end{aligned}
\end{equation}
\begin{equation}
\begin{aligned}
\label{eq:step2}
\mathcal{K}_2({w}',{h}',{r_1}) = \sum_{k_1=1}^{K_1} \sum_{k_2=1}^{K_2} \sum_{{r_2}=1}^{r^{(2)}} \mathcal{C}({r_1},{r_2},k_1,k_2)\mathcal{K}_1(w,h,{r_2}),
\end{aligned}
\end{equation}
\begin{equation}
\begin{aligned}
\label{eq:step3}
\mathcal{Y}({w}',{h}',f) = \sum_{{r_1}=1}^{r^{(1)}} M_1({r_1},f)\mathcal{K}_2({w}',{h}',{r_1}),
\end{aligned}
\end{equation}
where $\bmmc{K}_1 \in \mathbb{R}^{W\times H\times r^{(2)}}$ and $\bmmc{K}_2 \in \mathbb{R}^{{W}'\times {H}'\times r^{(1)}}$.

\begin{figure*}[ht]
\centering 
\includegraphics[width=0.8\linewidth]{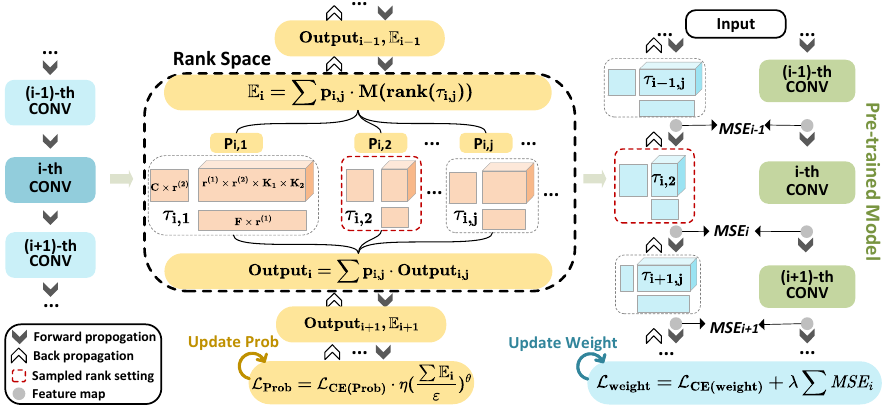} 
\vspace{-2mm}
\caption{The automatic rank selection process of HALOC via alternately updating rank selection probability and model weight.} 
\label{Fig.1}
\vspace{-5mm}
\end{figure*}

\section{Method}
\label{sec:method}



\subsection{Problem Formulation}
To realize practical and high-performance low-rank compression, we aim to automatically find the optimal ranks to minimize the accuracy loss and maximize the hardware performance (e.g., inference speed). Mathematically, for an $n$-layer convolutional neural network, this process can be formulated as a constrained optimization problem as below:
\begin{equation}
\begin{aligned}
&\min_{\{\bmmc{W}_i\}_{i=1}^{n}} \mathcal{L}(\{\bmmc{W}_i\}) \quad \textrm{s.t.} ~~ \sum_{i=1}^{n}\hbar(rank(\bmmc{W}_i)) \le \varepsilon,
\end{aligned}
\label{eq:opt_1}
\end{equation}
where $\{\bmmc{W}_i\}_{i=1}^{n}$ represents the weights of all layers of $n$-layer CNN model, $rank(\cdot)$ is the function that returns the tensor ranks of the factorized tensor cores, and $\varepsilon$ is the specific constraint for the practical hardware performance (e.g., latency). In addition, $\mathcal{L}(\cdot)$ and $\hbar(\cdot)$ denote the loss function and the layer-wise hardware performance, respectively.

\subsection{Design Challenges}
Solving the constrained problem described in Eq. \ref{eq:opt_1} is non-trivial but facing two main challenges. \ul{First}, because we cannot explicitly construct $rank(\cdot)$, the implicit mechanism that can automatically determine the rank setting is needed. To that end, some methods, such as approximation error-aware minimization and genetic algorithm, have been used to guide the search process. However, a common problem for the existing solutions is the insufficient exploration for the rank space. For instance, because the higher similarity between the original and the decomposed models does not necessarily mean higher accuracy, minimizing the approximation error will severely limit the exploration scope, causing unsatisfied compression performance. \ul{Second}, the search process of the state-of-the-art rank determination works cannot be extended to consider the hardware performance constraint described in Eq. \ref{eq:opt_1}. More specifically, the underlying mechanisms of the existing automatic rank selection methods, by their nature, can only support the differentiable constraint such as compression ratio; while the practical hardware performance, e.g., the measured latency, is non-differentiable, making it challenging to extend the prior solutions to the hardware-aware format.


\subsection{HALOC: Selecting Ranks as Architecture Search}

The above analyzed limitations of the existing rank selection methods call for more efficient solutions. To that end, we propose HALOC, a novel hardware-aware automatic low-rank DNN compression technique. HALOC is built on a key observation -- because low-rank compression aims to explore the structure-level redundancy of DNNs, the rank selection process essentially determines the architecture of the compressed models. Based on this perspective, \ul{automatic search for the suitable ranks can be interpreted as the automatic search for the proper low-rank structure,} opening up the opportunities of designing new rank selection solution guided by the philosophy of neural architecture search (NAS) \cite{cai2018proxylessnas,wu2019fbnet}.

Motivated by this hidden connection between setting the layer-wise ranks and searching the network architecture, we propose to develop efficient low-rank compression with NAS-inspired automatic rank selection. Our key idea is iteratively sampling and evaluating different candidate rank settings to learn the most suitable one in a differentiable way. More specifically, as illustrated in Figure \ref{Fig.1}, the HALOC framework consists of the following operations.

\textbf{Step-1. Constructing Low-Rank Search Space.} We first build an over-parameterized network $\bmmc{N}$ that consists of multiple candidate rank combinations. Notice that different from the case for NAS methods aiming to select from a group of candidate operators, the $n$-layer $\bmmc{N}(\mathcal{T}_{1}$, $\cdots, \mathcal{T}_{n})$ built for HALOC represents the ensemble of the candidate rank settings. More specifically, for the $i$-th layer of $\bmmc{N}$ as $\mathcal{T}_{i}$, it consists of $m_1m_2$ decomposition candidates as $\tau_{i,j}$ with rank setting $(r^{(1)}_{i,j_1}, r^{(2)}_{i,j_2})$, where $j_1=1,2,...,m_1$, $j_2=1,2,...,m_2$, and $j=1,2,...m_1m_2$. At the setup stage of automatic search, each $\tau_{i,j}$ is initialized via performing Tucker-2 decomposition on the $i$-th layer of the uncompressed model with rank set as $(r^{(1)}_{i,j_1}, r^{(2)}_{i,j_2})$. 

\textbf{Step-2. Updating Probabilities \& Weights.} Upon the construction and initialization of $\bmmc{N}$, we then alternately update the parameters of $\tau_{i,j}$ and the corresponding selection probability $p_{i,j}$ for the $i$-th layer. To be specific, because $p_{i,j}$ is calculated as $\bm{p}_i = Softmax(\bm{\alpha}_i)$, where $p_{i,j} \in \bm{p}_i$ and $\bm{\alpha}_i$ is an learnable vector, the update of all $p_{i,j}$'s can be simultaneously performed via using the backward propagation on the validation dataset. On the other hand, HALOC updates the weights of $\tau_{i,j}$ in a selective way. To reduce the computational cost, guided by the selection probability $p_{i,j}$, each time only one $\tau_{i,j}$ is sampled and updated per layer, and this weight update process is based on the loss function defined on the training dataset. After multiple iterations of alternated update for probabilities and weights, the final selected decomposed candidate for the $i$-th layer is the $\tau_{i,j}$ with the largest $p_{i,j}$.


\ul{\textbf{Considering Hardware Constraints.}} As outlined in Eq. \ref{eq:opt_1}, the constraints on the hardware performance should be taken into account when designing low-rank compression technique towards practical applications. Unlike the existing automatic low-rank compression works that cannot properly include hardware performance into the design phase, HALOC enjoys an attractive benefit of its inherent compatibility for hardware-awareness. More specifically, considering the hardware performance (e.g., latency) is non-differentiable, we use a prediction model to estimate the expected hardware performance of the $i$-th layer as follows:
\begin{equation}
\begin{aligned}
&\mathbb{E}_i=\sum_{j=1}^{m_1m_2} p_{i,j} \cdot M(rank(\tau_{i,j})), \\
\end{aligned}
\label{eq:nas_1}
\end{equation}
where $M(\cdot)$ denotes the \textit{layer-wise} hardware performance for the decomposition candidate $\tau_{i,j}$ with rank setting $(r^{(1)}_{i,j_1}, r^{(2)}_{i,j_2})$. Here the hardware performance for one candidate can be either pre-measured from the target computing devices or estimated from a regression model. Essentially, the layer-wise prediction model described in Eq. \ref{eq:nas_1} can be viewed as the weighted sum of the performance of the decomposition candidates for the current layer, and it is then integrated to the process for updating the selection probabilities as follows:
\begin{equation}
\begin{aligned}
&\mathcal{L}_{Prob} = \mathcal{L}_{CE(Prob)} \cdot \eta (\frac{\sum_{i=1}^{n}\mathbb{E}_i}{\varepsilon})^\theta,\\
\end{aligned}
\label{eq:nas_2}
\end{equation}
where $\mathcal{L}_{CE(Prob)}$ is the cross-entropy loss on the validation dataset, and $\eta$ and $\theta$ are the hyperparameters that adjust the impact of hardware constraints on the overall search process \cite{wu2019fbnet}. 

\subsection{Questions to be Answered}

As outlined above, the HALOC framework can be developed from the perspective of architecture search. However, as we will further analyze in this subsection, because HALOC aims to perform automatic low-rank compression, a task that is essentially different from NAS, it is facing several unique design challenges when searching in the rank space. To address these issues and realize automatic rank selection efficiently, next we explore to answer two important questions.

\ul{\textbf{Question 1:}} \textit{How should we set the proper search scope to realize sufficient exploration in the rank space with affordable search cost?}

\ul{Analysis.} A very key challenge for HALOC is the extremely huge search space, which is much larger than the scope explored in the existing NAS works. For instance, there are only $1.5\times10^{17}$ potential network architectures when searching an 18-layer CNN using NAS \cite{wu2019fbnet}; while there exist $2.2\times 10^{71}$ candidates of rank combination when performing low-rank compression on a ResNet-18 model. In other words, \ul{the rank space for low-rank compression is much larger than the architecture space for NAS}, hindering the automatic rank selection in a timely and efficient way. 

Essentially, the ultra-large search space of HALOC results from two sources. \ul{First}, for each layer, the number of rank candidates is inherently much more than that of operator candidates. For instance, the uncompressed layer (layer3.1.conv1) in ResNet-18 model has full rank size as 256, meaning that there exist at least 256 possible ranks for $r^{(1)}$ or $r^{(2)}$ can be selected for low-rank compression; while the operator for building one layer are only considered and selected from a limited set, e.g., around 10 candidates as indicated in  \cite{liu2018darts,cai2018proxylessnas,wu2019fbnet}. \ul{Second}, because it is very common that multiple rank modes are needed in low-rank compression, e.g., a Tucker-2 decomposed layer is determined by two ranks ($r^{(1)}$ and $r^{(2)}$ in Eq. \ref{eq:tucker2}), the corresponding combinatorial effect further drastically enlarges the overall to-be-explored rank space.



\begin{figure}
  \begin{subfigure}{\linewidth}
    \centering
    \includegraphics[width=0.75\linewidth]{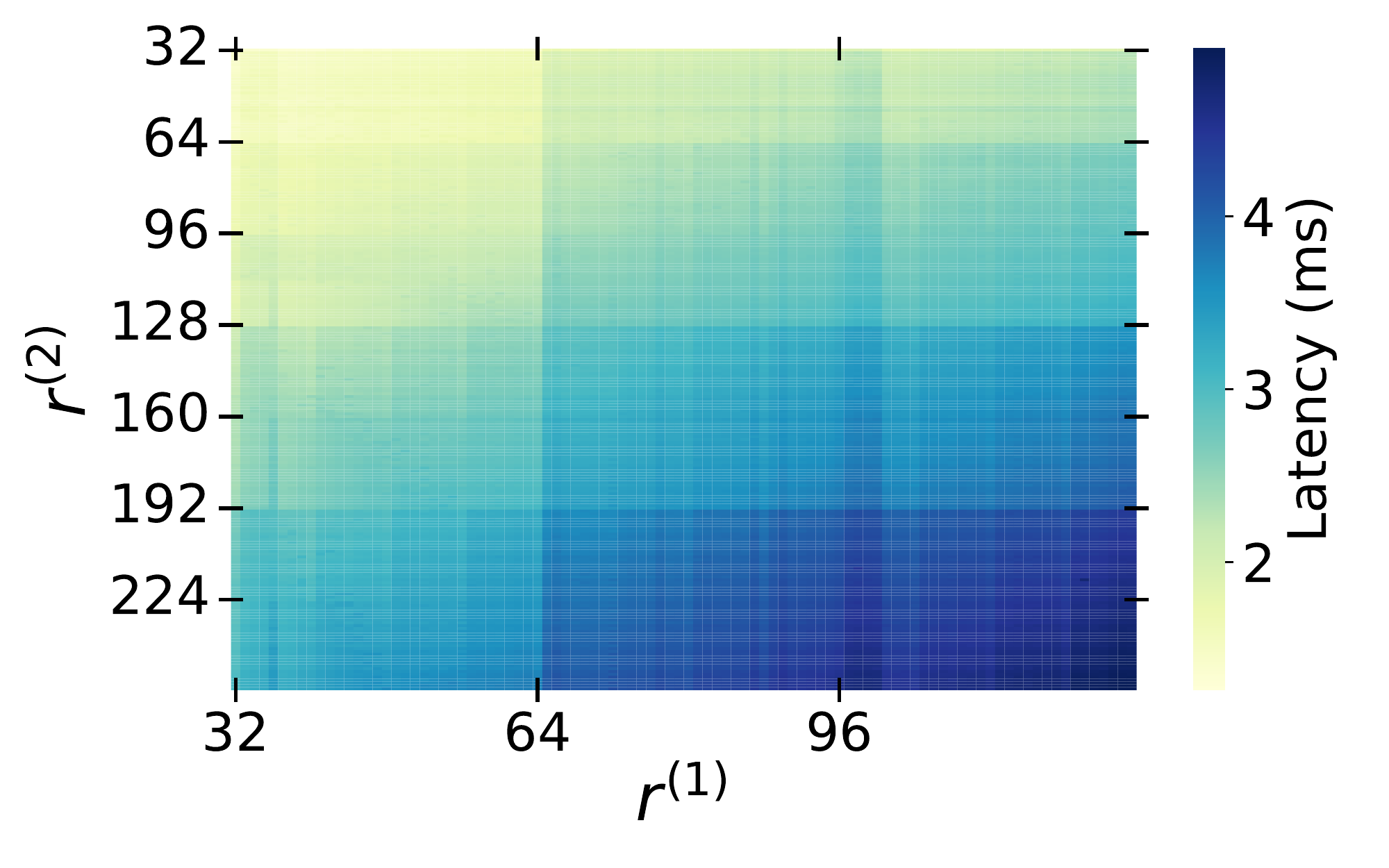}
    \vspace{-2mm}
    \caption{Layer3.0.conv on Nvidia RTX 2080. Batch size is 128.}
    \label{sf1}
  \end{subfigure}
  \begin{subfigure}{\linewidth}
    \centering
    \includegraphics[width=0.75\linewidth]{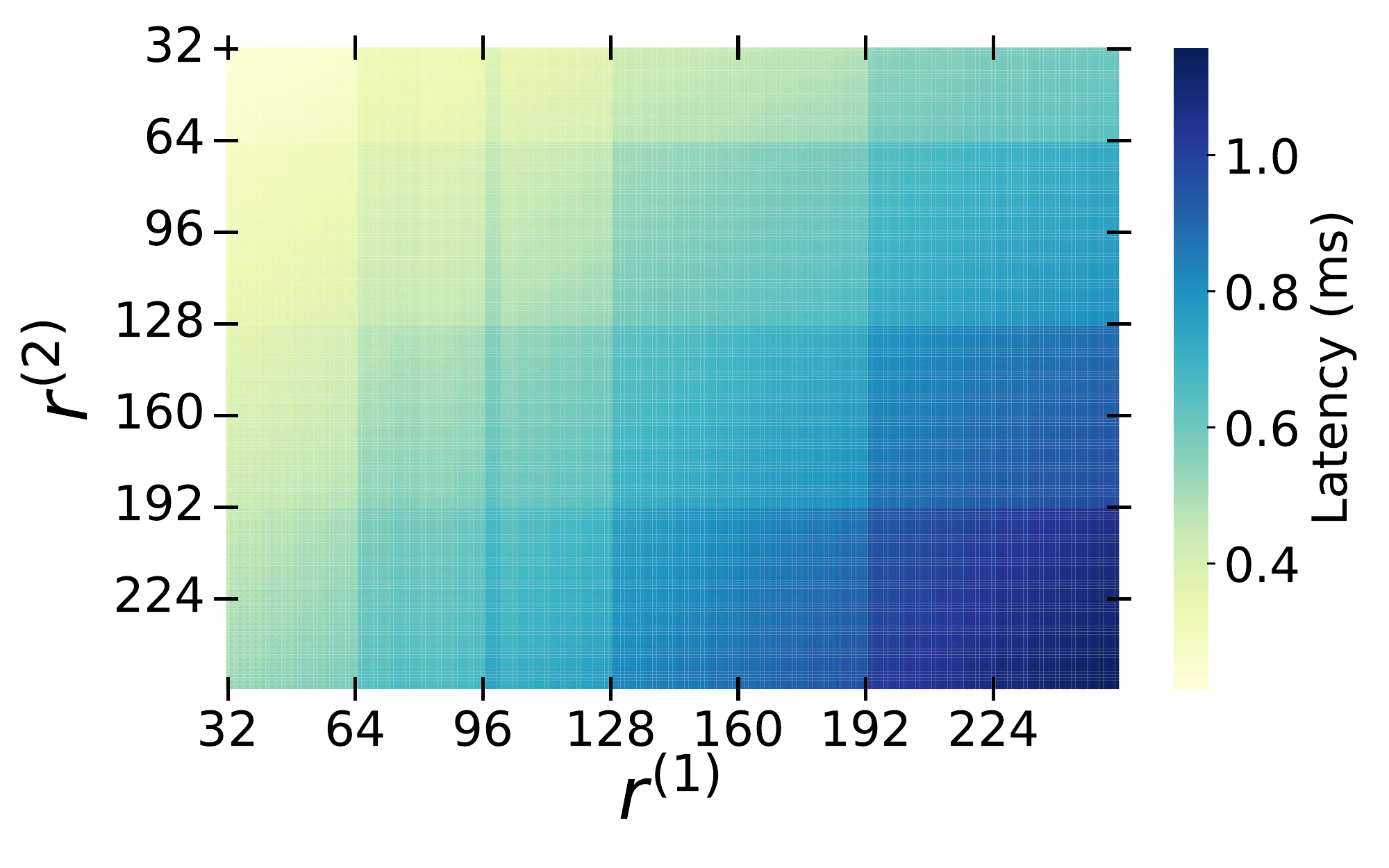}
    \vspace{-2mm}
    \caption{Layer3.0.conv2 on Nvidia Tesla V100. Batch size is 128.}
    \label{sf2}
  \end{subfigure}
  \vspace{-2mm}
  \caption{The heat map of the measured latency on the practical hardware for Tucker-2 format layers of ResNet-18 with different rank settings. Batch processing is used for stable measurement.}
  \label{fig:latency}
\end{figure}


\ul{Our Proposal.} Based on the above analysis, we propose to perform efficient search in the rank space to achieve good balance between the efficiency and globality of the exploration. \ul{First}, we reduce the numbers of the rank candidates in a hardware-aware way. More specifically, we analyze the impact of different rank settings of single layer on the measured latency, and we discover that many ranks, though corresponding to different FLOPs, bring very similar latency on the hardware platforms. For instance, as illustrated in Figure \ref{fig:latency}, when performing Tucker-2 decomposition on a convolutional layer with different rank settings, obvious latency change can only be observed when $r^{(1)}$ and $r^{(2)}$ increase by 32 or 64 -- a phenomenon that exists across different types of computing platforms. \ul{We hypothesize that the most probable reason causing this phenomenon is the under-utilization of hardware resource for some rank settings.} As analyzed in \cite{qin2020sigma}, when the size of the (decomposed) layer cannot match the size of computing resource (e.g., arrays of processing elements in ASIC accelerator and 32-thread wrap in desktop GPU) very well, many computing units would be idle and under-utilized, causing that the execution of the two low-rank layers with different ranks consumes the same clock cycles. Motivated by this discovery, we propose the following design principle to only perform search on the rank values that correspond to obvious latency changes:

\vspace{1.5mm}
\textit{\textbf{Design Principle-1:} To make good balance between search cost and rank granularity, the rank candidates in HALOC is set as the multiples of a constant (typically 32).} 

\vspace{1.5mm}
Notice that when the number of input/output channels of a convolutional layer is small (e.g., 64 or 32), the suggested constant can be set as 16 or 8 to ensure the sufficient coverage for searching in the rank space. \ul{More importantly, when the target compressed model size/FLOP is already pre-known, the setting for this constant can be automatically calculated according to the pre-set compression ratio. Please refer to \textbf{Appendix} for more details.}

\ul{Second}, in addition to specify the search granularity for $r^{(1)}$ and $r^{(2)}$, we also identify their suitable relationship to further reduce the search space. Based on our in-depth analysis (\ul{details described in \textbf{Appendix}}), we propose the following design principle:

\vspace{1.5mm}
\textit{\textbf{Design Principle-2:} For a Tucker-2-format layer, equal rank setting ($r^{(1)}=r^{(2)}$) can be adopted to simplify the rank search process with good approximation performance.}

\begin{figure}
    \centering
    \includegraphics[width=0.75\linewidth]{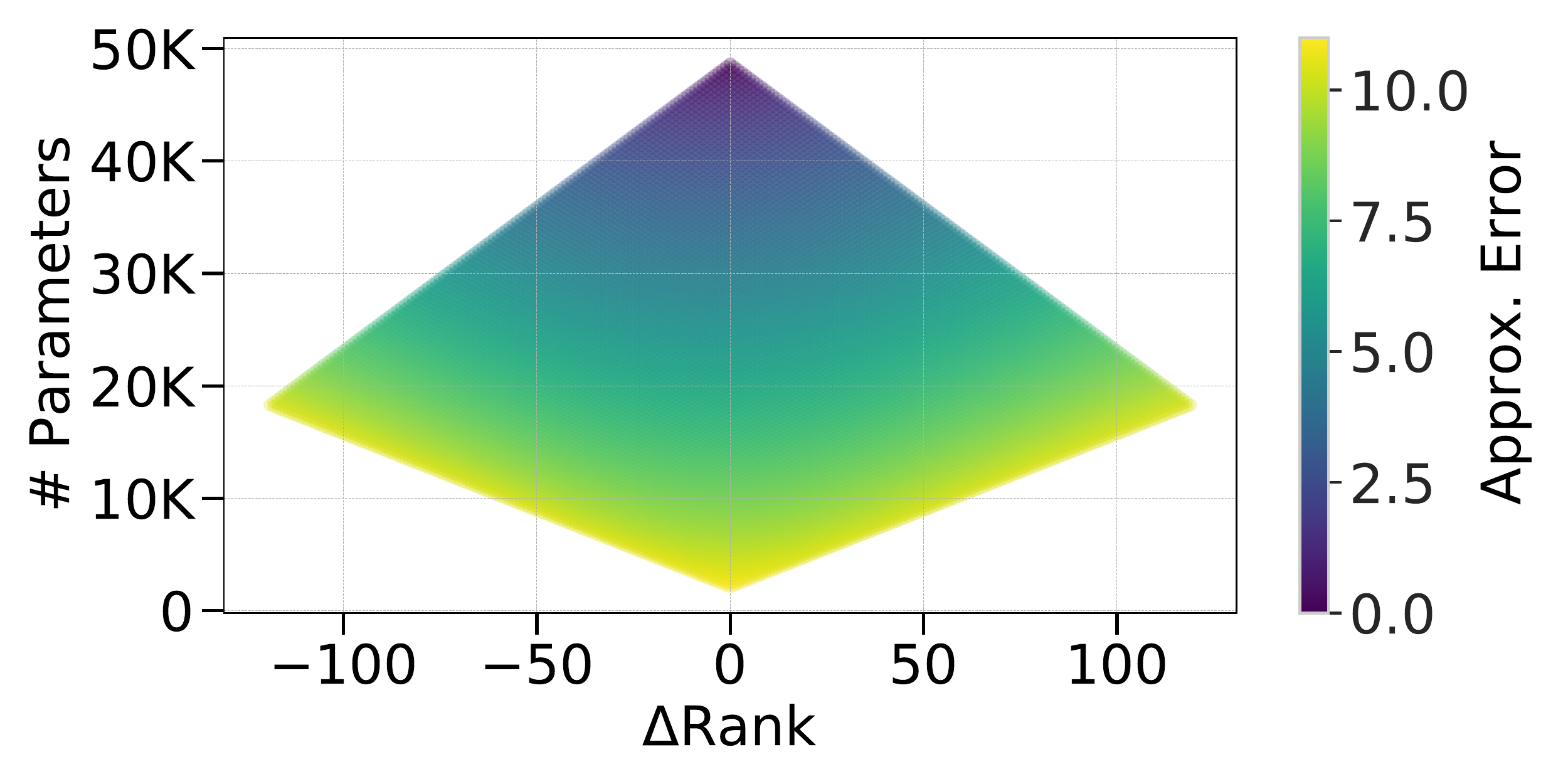}
    \vspace{-3mm}
    \caption{Approximation error of a decomposed  convolutional layer (layer2.1.conv2) in ResNet-18 model with different rank discrepancies ($\Delta Rank=r^{(1)}-r^{(2)}$) and target numbers of parameters after decomposition.}
    \label{fig:rank_appr}
    \vspace{-3mm}
\end{figure}

\vspace{1.5mm}
Figure \ref{fig:rank_appr} empirically illustrates the approximation errors with different rank discrepancy ($\Delta Rank=r^{(1)}-r^{(2)}$) given the same number of weights for a decomposed layer. It is seen that setting $r^{(1)}$ equal to $r^{(2)}$ indeed brings much smaller approximation error than most of other configurations, leading to the good balance between the simplicity and quality of rank search. Overall, guided by these design principles, when using HALOC to compress ResNet-18 model, the number of the possible rank combinations can be reduced from $2.2\times10^{71}$ to $7.8\times10^{13}$, leading to even smaller search space than that of the existing NAS methods.


\ul{\textbf{Question \#2:}} \textit{What is the proper scheme to mitigate the interference between different selected rank settings?}

\ul{Analysis.} Another issue that may affect the performance of HALOC is the potential interference between different low-rank structure candidates in the search process. To be specific, because a decomposition candidate $\tau_{i,j}$ in the $i$-th layer may be connected to different $\tau_{i+1,j}$'s of the ($i+1$)-th layer in different sampling trials, the discrepancy between the gradient directions of $\tau_{i+1,j_1}$ and $\tau_{i+1,j_2}$ imposed on  $\tau_{i,j}$, if large, would bring interference on the weight/probabilities update process, causing insufficient training and/or slow convergence. Notice that similar phenomenon has also been observed in the weight-sharing NAS \cite{shu2019understanding,zhang2020deeper}.


\ul{Our Proposal.} To address this challenge, we propose to use the prior information of the uncompressed model to mitigate the potential interference. \ul{Our key idea lies in the hypothesis that the higher similarity between $\tau_{i+1,j_1}$ and $\tau_{i+1,j_2}$ renders higher similarity between their gradients directions imposed on $\tau_{i,j}$.} Therefore, considering each  $\tau_{i+1,j}$ can be roughly interpreted as a type of approximation for the original uncompressed weight tensor $\bmmc{W}_{i+1}$ in the ($i+1$)-th layer, we propose to use the pre-known information of $\bmmc{W}_{i+1}$ to improve the similarity between $\tau_{i+1,j_1}$ and $\tau_{i+1,j_2}$. To be specific, we can regularize the search process to force both $\tau_{i+1,j_1}$ and $\tau_{i+1,j_2}$ to approach $\bmmc{W}_{i+1}$, making them become more similar. 

In general, this uncompressed layer-approaching strategy can be realized via two ways: we can make either the weights or the output feature maps of $\tau_{i+1,j}$ approach those of $\bmmc{W}_{i+1}$. HALOC adopts the feature map-based approaching solution because of two reasons. \ul{\textbf{1) Low computational costs.}} To measure the similarity between the weights of $\tau_{i+1,j}$ and $\bmmc{W}_{i+1}$, we need to reconstruct the approximated $\bmmc{W}_{i+1}$ from $\tau_{i+1,j}$ via using Eq. \ref{eq:tucker2}. Unfortunately, such reconstruction operation is not well supported by the current version of PyTorch/Tensorflow, causing very slow execution speed. On the other hand, calculating the output feature maps of $\bmmc{W}_{i+1}$ is equivalent to performing forward prorogation for the original model, an execution that is much faster and well supported by GPU acceleration. \ul{\textbf{2) Rich information.}} As indicated in \cite{lin2020hrank} the feature maps naturally capture and contain the rich and critical information for both model and data. Compared with weight-approaching strategy that only reflects static model characteristics, approaching feature maps can provide additional dynamic data-aware alignment. We observe that such feature map-preferred philosophy is also advocated and adopted in a set of channel pruning works \cite{hou2022chex,NEURIPS2021_ce6babd0,tang2020scop,lin2020hrank}, demonstrating the importance and usefulness of feature map information. Therefore, we use the following mean square error (MSE)-based loss to measure the feature map similarity:
\begin{equation}
\begin{aligned}
\mathcal{L}_{approach} = \sum_{i=1}^{n} MSE(Fmap_{decomp,i}, Fmap_{org,i}) ,
\end{aligned}
\label{eq:min_interfere}
\end{equation}
where $Fmap_{org,i}$ and $Fmap_{decomp,i}$ denote the feature maps of the original weight tensor and the sampled $\tau_{i,j}$ in the $i$-th layer, respectively. Then the loss function for updating the weights is as follows:
\begin{equation}
\begin{aligned}
\mathcal{L}_{weight} = \mathcal{L}_{CE(weight)} + \lambda \mathcal{L}_{approach},
\end{aligned}
\label{eq:child_loss}
\end{equation}
\textcolor{black}{
where $\mathcal{L}_{CE(weight)}$ is the cross-entropy loss defined on the training dataset, and $\lambda$ is the scaling parameter that controls the impact of feature map-approaching loss.}

\section{Experiments}
We evaluate the performance of HALOC for compressing different CNN models on CIFAR10 \cite{krizhevsky2009learning}, and ImageNet \cite{deng2009imagenet} datasets. We also measure the practical latency of the automatic compressed models on different types of hardware platforms, including desktop GPU, embedded GPU and ASIC accelerator.

\textbf{Training Details.} We use the standard SGD optimizer with Nesterov momentum as 0.9 for model training. The learning rates are initilized as 0.1 and 0.01 for CIFAR-10 and ImageNet, respectively, and they are then scaled down by 0.2 every 55 epochs. In addition, batch size and weight decay are set as 256 and 0.0001, respectively.

\subsection{CIFAR-10 Results}
Table \ref{tbl:acc-cifar10} shows the evaluation results for compressing different CNN models on CIFAR-10 dataset. For compressing ResNet-20 model, our method can achieve even 0.07\% higher top-1 accuracy than the baseline model with 72.6\% FLOPs reduction and 76.1\% model size reduction. Compared with ALDS \cite{liebenwein2021compressing}, the state-of-the-art automatic low-rank compression work, HALOC achieves 0.35\% accuracy increase with higher computational and memory cost reductions, demonstrating the outstanding compression performance of our proposed approach.

\begin{table}[H]
\small
\setlength\tabcolsep{3pt}
\begin{center}
\begin{tabular}{llcccccc}
\toprule
 &{\multirow{2}{*}{\textbf{Method}}} &{\multirow{2}{*}{{\makecell{\textbf{Comp.}\\\textbf{Type}}}}} &{\multirow{2}{*}{\makecell{\textbf{Auto.}\\\textbf{Rank}}}} &{\multirow{2}{*}{\makecell{\textbf{Top-1}\\\textbf{(\%)}}}} &{\multirow{2}{*}{\makecell{\textbf{FLOPs}\\\textbf{($\downarrow$\%)}}}}& {\multirow{2}{*}{\makecell{\textbf{Params.}\\\textbf{($\downarrow$\%)}}}}\\\\
\toprule
& ResNet-20  & Baseline      & -    & 91.25   & -          & -    \\
\midrule
 & \textbf{HALOC} & Low-rank &\cmark  & \textbf{91.32} & \textbf{72.20}  & \textbf{76.10}         \\
 & ALDS      & Low-rank &\cmark  & 90.92          & 67.86           & 74.91      \\
 & LCNN  &Low-rank  &\cmark  &90.13  &66.78      &65.38  \\ 
 & PSTR-S    & Low-rank &\cmark  & 90.80           & -   & 60.87      \\ 
 & Std. Tucker  & Low-rank  &\xmark  &87.41  &-      &61.54  \\
\toprule
 &VGG-16      & Baseline      & -   &92.78   & -          & - \\
\midrule
 & \textbf{HALOC}  & Low-rank  &\cmark     & \textbf{93.16} & \textbf{86.44}  & \textbf{98.56}         \\
 & ALDS  & Low-rank     &\cmark       & 92.67          & 86.23           & 95.77      \\
 & LCNN  &Low-rank  &\cmark  &92.72  &85.47      &91.14  \\
 & DECORE  & Pruning     &-       & 92.44          & 81.50           & 96.60      \\ 
 & Spike-Thrift  & Pruning  &-       & 91.79          & -           & 97.01    \\ 
\bottomrule
\end{tabular}
\caption{Comparison with different compression approaches for ResNet-20 and VGG-16 on CIFAR-10. ALDS \cite{liebenwein2021compressing},
LCNN \cite{idelbayev2020low},
PSTR-S \cite{li2022heuristic},
Std. Tucker \cite{kim2016compression},
DECORE \cite{Alwani_2022_CVPR},
Spike-Thrift \cite{Kundu_2021_WACV}.
}
\label{tbl:acc-cifar10}
\vspace{-4.5mm}
\end{center}
\end{table}

\begin{table}[H]
\small
\setlength\tabcolsep{1.3pt}
\begin{center}
\vspace{-2mm}
\begin{tabular}{llcccccc}
\toprule
 &{\multirow{2}{*}{\textbf{Method}}} &{\multirow{2}{*}{\makecell{\textbf{Comp.}\\\textbf{Type}}}} &{\multirow{2}{*}{\makecell{\textbf{Auto.}\\\textbf{Rank}}}} &{\multirow{2}{*}{\makecell{\textbf{Top-1}\\\textbf{(\%)}}}}&{\multirow{2}{*}{\makecell{\textbf{Top-5}\\\textbf{(\%)}}}}
 &{\multirow{2}{*}{\makecell{\textbf{FLOPs}\\\textbf{($\downarrow$\%)}}}}
 & {\multirow{2}{*}{\makecell{\textbf{Params.}\\\textbf{($\downarrow$\%)}}}}\\\\
\toprule
&ResNet-18  & Baseline   &-      &69.75   &89.08     &-      &- \\
\midrule
 & \textbf{HALOC} & Low-rank   &\cmark  & \textbf{70.65} & \textbf{89.42} & \textbf{66.16} & 63.64 \\
 & \textbf{HALOC} & Low-rank   &\cmark & 70.14    & 89.38     & 63.81     & \textbf{71.31} \\
 & ALDS    & Low-rank    &\cmark      & 69.22          & 89.03          & 43.51          & 66.70           \\ 
 & TETD      & Low-rank     &\xmark     &-           &89.08            & 59.51             & -    \\ 
 & Stable EPC      & Low-rank    &\cmark    &-           &89.08            & 59.51             & -    \\ 
 & MUSCO    & Low-rank  &\xmark     & 69.29          & 88.78          & 58.67          & -  \\ 
 & CHEX  &Pruning  &- & 69.60          & -          & 43.38          & -           \\ 
 & EE     & Pruning  &-     & 68.27          & 88.44          & 46.60          & -           \\ 
 & SCOP  & Pruning &-  & 69.18          & 88.89          & 38.80          & 39.30           \\ 
\toprule
&\multicolumn{2}{l}{MobileNetV2~~Baseline}  &- &71.85   &90.33     &-      &-\\

\midrule
 & \textbf{HALOC} & Low-rank &\cmark & \textbf{70.98} & \textbf{89.77} & 24.84 &40.03 \\
 & \textbf{HALOC} & Low-rank &\cmark &66.37  &87.02  & \textbf{45.65} & \textbf{62.59} \\
 & ALDS     & Low-rank   &\cmark & 70.32          & 89.60       & 11.01          & 32.97\\ 
 & HOSA     & Pruning   &- &64.43           & -       & 43.65          & 27.13     \\ 
 & DCP     & Pruning   &- &64.22           & -       & 44.75          & 25.93      \\ 
 & FT     &Pruning   &-  & 70.12          & 89.48       & 20.23          & 21.31  \\ 
 
\bottomrule
\end{tabular}
\caption{Comparison with different compression approaches for ResNet-18 and MobileNetV2 on ImageNet. ALDS \cite{liebenwein2021compressing},
TETD \cite{Yin_2021_CVPR},
Stable EPC \cite{phan2020stable},
MUSCO \cite{gusak2019automated},
CHEX \cite{hou2022chex},
EE \cite{Zhang_2021_ICCV},
SCOP \cite{tang2020scop},
HOSA \cite{chatzikonstantinou2020neural},
DCP \cite{NEURIPS2018_55a7cf9c},
FT \cite{he2017channel}.  
}
\label{tbl:acc-imagenet}
\vspace{-6mm}
\end{center}
\end{table}

For compressing VGG-16 model, HALOC also shows high compression performance. It brings 0.38\% higher accuracy than the original uncompressed model with 86.44\% FLOPs reduction and 98.56\% model size reduction. Compared with the state-of-the-art pruning and low-rank compression works, HALOC consistently achieves higher accuracy with more aggressive compression efforts.

\begin{table*}[tb]
\vspace{-8mm}
\begin{center}
\begin{tabular}{cccccccccc}
\toprule
  \multicolumn{1}{c}{\multirow{4}{*}{\textbf{Hardware}}} & \multirow{4}{*}{\makecell{\textbf{Method}}} &\multicolumn{4}{c}{\textbf{ResNet-18}} &\multicolumn{4}{c}{\textbf{MobileNetV2}} \\ \cmidrule(lr){3-6} \cmidrule(lr){7-10}
 &&\multirow{2}{*}{\makecell{\textbf{Top-1} \\ \textbf{(\%)} }} &\multirow{2}{*}{\makecell{\textbf{Top-5}\\ \textbf{(\%)} }} & \multirow{2}{*}{\makecell{\textbf{FLOPs}\\ \textbf{(M)}}}  & \multirow{2}{*}{\makecell{\textbf{Throughput}\\ \textbf{(images/s)}}}  &\multirow{2}{*}{\makecell{\textbf{Top-1} \\ \textbf{(\%)}}} &\multirow{2}{*}{\makecell{\textbf{Top-5}\\ \textbf{(\%)} }} & \multirow{2}{*}{\makecell{\textbf{FLOPs}\\ \textbf{(M)}}}  & \multirow{2}{*}{\makecell{\textbf{Throughput}\\ \textbf{(images/s)}}}\\\\
\toprule
\multicolumn{1}{c}{\multirow{2}{*}{\makecell{NVIDIA\\ Tesla V100}}}
& Original    &69.75   &89.08       & 1819.07  & 4362.1    &71.85   &90.33     & 314.19   & 3877.3 \\
&   \textbf{HALOC}   & 69.75  & 88.93  & \textbf{553.13}   & \textbf{6360.5}    & 70.86  & 89.77        & \textbf{245.52}   & \textbf{3993.6}\\
\midrule
\multicolumn{1}{c}{\multirow{2}{*}{\makecell{NVIDIA\\ Jetson TX2}}}
& Original    &69.75   &89.08   & 1819.07  & 86.3    &71.85   &90.33      & 314.19   & 112.1 \\
&   \textbf{HALOC}   & 70.14  & 89.38  & \textbf{658.26}   & \textbf{151.0}    & 70.80   & 89.55  & \textbf{240.99}   & \textbf{117.0}\\
\midrule
\multicolumn{1}{c}{\multirow{2}{*}{\makecell{ASIC\\ Eyeriss}} }    
& Original    &69.75   &89.08       & 1819.07  & 121.4     &71.85   &90.33      & 314.19   &496.3\\
&   \textbf{HALOC}   & 70.65  & 89.42  & \textbf{615.62}   & \textbf{247.0}       & 70.83  & 89.65        & \textbf{229.13}   & \textbf{590.2}\\   
\bottomrule  
\end{tabular}
\vspace{-2mm}
\caption{Measured Speedup for compressed ResNet-18 and MobileNetV2 on different computing platforms. Hardware-aware automatic rank selection is adopted in the low-rank compression process.}
\label{tbl:perf}
\vspace{-6mm}
\end{center}
\end{table*}

\subsection{ImageNet Results}
Table \ref{tbl:acc-imagenet} shows the performance of different compression methods on ImageNet dataset. For compressing ResNet-18 model, HALOC achieves 0.9\% higher top-1 accuracy than the baseline model with 66.16\% FLOPs reduction and 63.64\% model size reduction. Compared with the state-of-the-art automatic low-rank compression method ALDS, HALOC enjoys 1.4\% higher top-1 accuracy with much lower computational costs. Compared with the state-of-the-art pruning work CHEX \cite{hou2022chex}, HALOC also achieves 0.54\% accuracy increase with much fewer FLOPs. 

Besides, our approach also shows impressive performance for compressing MobileNetV2, a task that is conventionally challenging for low-rank compression approach.  From Table \ref{tbl:acc-imagenet} it is seen that HALOC can achieve higher performance than the existing low-rank compression and pruning works. In particular, compared with ALDS, the method that reports the best performance of compressing MobileNetV2 among all the existing low-rank solutions, HALOC shows 0.66\% top-1 accuracy increase with higher FLOPs reduction and model size reduction.


\subsection{Practical Speedups on Hardware Platforms}
We also measure the practical speedups brought by our hardware-aware solution on various computing hardware, including Nvidia Tesla V100, Nvidia Jetson TX2, and ASIC accelerator Eyeriss \cite{chen2016eyeriss}. Here the performance of Eyeriss is reported via using Timeloop \cite{parashar2019timeloop} with 45nm CMOS technology setting. As shown in Table \ref{tbl:perf}, the ResNet-18 and MobileNetV2 models that are compressed by HALOC shows considerable measured speedups across different platforms, demonstrating the practical effectiveness of our proposed hardware-aware low-rank compression solution.


\begin{figure}[ht]
  \begin{subfigure}{\linewidth}
    \centering
    \caption{The curve of training loss ($\mathcal{L}_{weight}$)}
    \vspace{-2mm}
    \includegraphics[width=\linewidth]{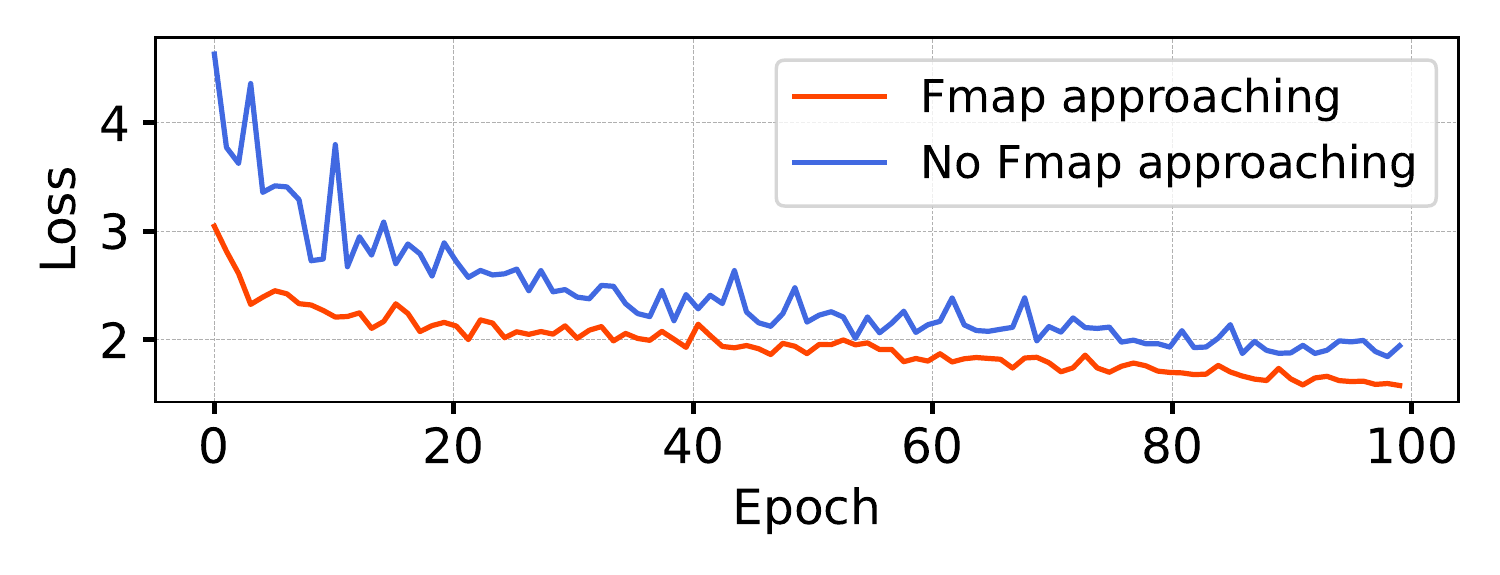}
    \label{sf:train_loss_curve}
    \vspace{-4mm}
  \end{subfigure}
  \begin{subfigure}{\linewidth}
    \centering
    \caption{The change of the model size in the rank search process}
    \vspace{-2mm}
    \includegraphics[width=\linewidth]{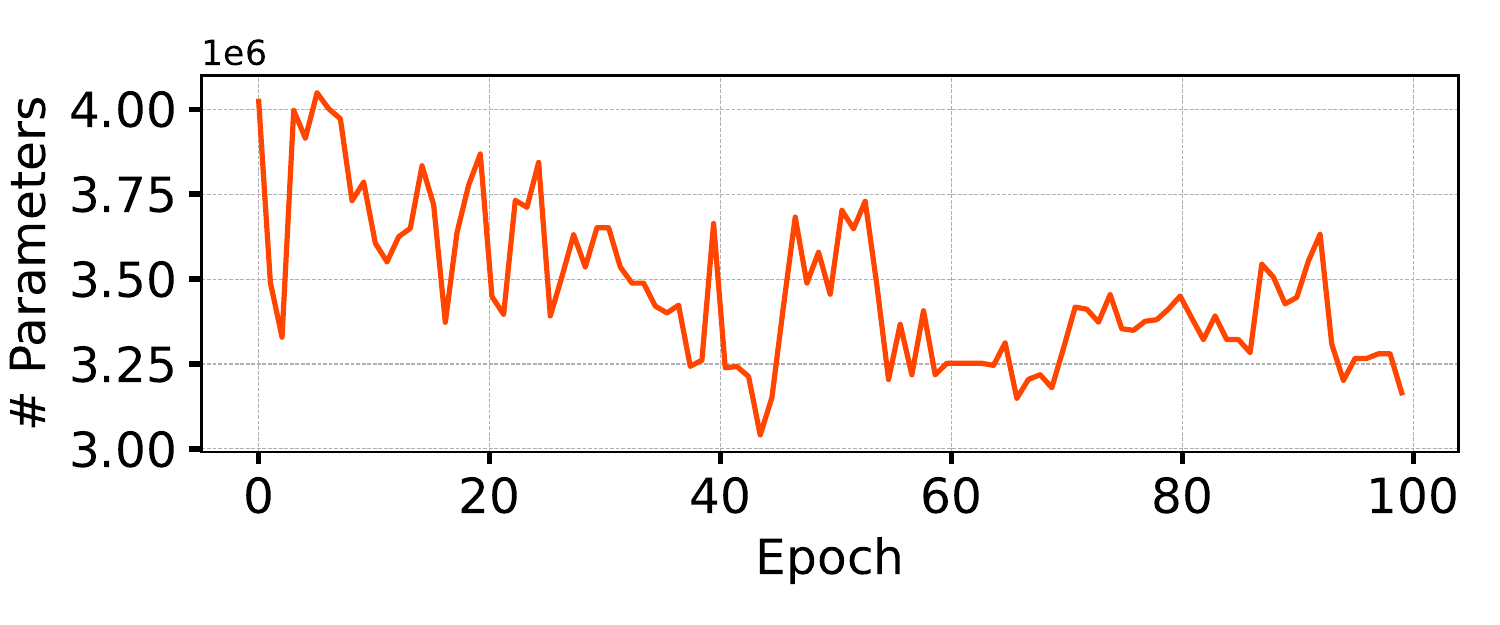}
    \label{sf:params_variation}
    \vspace{-5mm}
  \end{subfigure}
   \begin{subfigure}{\linewidth}
    \centering
    \caption{The final rank distribution after automatic rank search}
    \vspace{-1mm}
    \includegraphics[width=\linewidth]{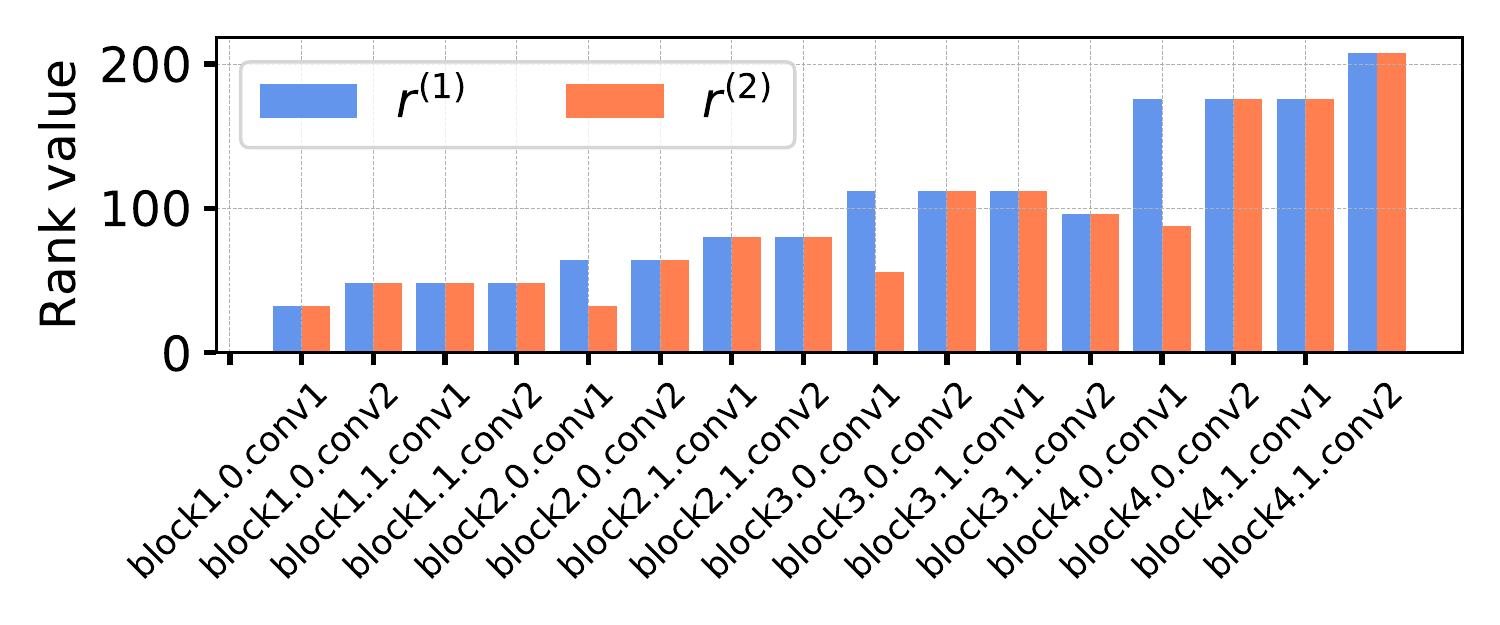}
    \label{sf:rank_dis}
  \end{subfigure}
  \vspace{-9mm}
  \caption{Experimental analysis for compressing ResNet-18 on ImageNet dataset using HALOC.}
  \label{fig:curves}
  \vspace{-6mm}
\end{figure}

\subsection{Analysis \& Discussion}
To obtain in-depth understanding and verify the effectiveness of HALOC, we perform some ablation study and experimental analysis for compressing ResNet-18 model on ImageNet dataset. First, we examine the impact of the proposed feature map-approaching strategy on the search process. As shown in Figure \ref{fig:curves} (a), the use of feature map-based regularization brings significant lower training loss, which further translates to higher accuracy for the final compressed model. In addition, Figure \ref{fig:curves} (b) shows the change of the total number of parameters during the search process. It is seen that HALOC indeed dynamically explores the rank space towards reducing the overall model complexity. The final rank distribution of the automatically compressed ResNet-18 model is visualized in Figure \ref{fig:curves} (c).

\section{Conclusion}
This paper proposes HALOC, a hardware-aware automatic low-rank compression framework for compact DNN models. By interpreting rank selection as architecture search process, HALOC enables efficient automatic rank determination in a differentiable way, realizing high-performance low-rank compression. Experimental results on both algorithm and hardware aspects demonstrate the effectiveness of our proposed approach.

{
\bibliography{reference}
}
\clearpage
\part*{ Appendix }
\textbf{Rank Space with Pre-set Compression Ratio.} \textit{Design Principle-1} of the main manuscript discusses the suggested empirical setting of the rank granularity. Here we present a more analytic solution for a special case -- when the target model compression ratio is pre-known. Our key idea is to first use the overall compression ratio to obtain a relaxed range for the estimated layer-wise rank value reduction (as compared to full rank), which can be then used to construct the rank space, and then we leverage the rank search process described in the main manuscript to identify the accurate rank setting that satisfies with the compression ratio requirement.\\

More specifically, suppose the overall target compression ratio is $\alpha$, then the average compression ratio of each layer is also first roughly set as $\alpha$. For a CONV layer with weight tensor $\mathcal{W} \in \mathbb{R}^{F \times C \times K_1 \times K_2}$, the number of parameters for original CONV and Tucker-2 Decomposed CONV can be calculated as $n_{org}=FCK_1K_2$ and $n_{tucker}=Fr^{(1)}+Cr^{(2)}+r^{(1)}r^{(2)}K_1K_2$, respectively. Then we have
\begin{equation}
\begin{aligned}
&\left\{
\begin{aligned}
&n_{tucker} = \frac{C^2}{\alpha_{rank}} + \frac{F^2}{\alpha_{rank}} + \frac{C^2F^2}{\alpha^2_{rank}}\\
&\frac{n_{org}}{\alpha} = \frac{FCK_1K_2}{\alpha} = n_{tucker}\\
\end{aligned}
\right.
\\& \alpha_{rank} = \frac{2FCK_1K_2}{\sqrt{(C^2+F^2)^2+\frac{4(FCK_1K_2)^2}{\alpha}}-C^2-F^2},
\end{aligned}
\label{eq:r_comp}
\end{equation}
where $\alpha_{rank}$ is a scaling parameter which represents the reduction of the assigned ranks from the original full rank value, $r^{(1)}=\frac{F}{\alpha_{rank}}$ and $r^{(2)}=\frac{C}{\alpha_{rank}}$. Then we can use Algorithm \ref{alg:algorithm} to determine the search rank space for Tucker-2 decomposition.

\textbf{Determining Discrepancy between $r^{(1)}$ and $r^{(2)}$.}
To explore the best suitable $|\Delta Rank=r^{(1)}-r^{(2)}|$, we aim to minimize the approximation error given the  same number of post-decomposition parameters. 
According to the Higher-Order Singular Value Decomposition (HOSVD) \cite{hitchcock1928multiple,de2000multilinear}, we use the percentage of the sum of the truncated singular values to evaluate the degree of information preservation after decomposition as follows:
\begin{equation}
\begin{aligned}
\rho  = \frac{\sum_{i=1}^{r^{(1)}} s_{1,i}}{\sum_{i=1}^{F} s_{1,i}}  \cdot \frac{\sum_{j=1}^{r^{(2)}} s_{2,j}}{\sum_{j=1}^{C} s_{2,j}},
\end{aligned}
\label{eq:pca_ratio}
\end{equation}
where $s_{1,i}$ and $s_{2,j}$ indicate the singular values, and $\rho$ is the percentage of the sum of the truncated singular values. Without loss of generality, we simplify the calculation by assuming that the $s_{1,i}=s_{2,j}$, then we have
\begin{equation}
\label{eq:eq3}
\begin{aligned}
\rho = \frac{r^{(1)}s_{1,1}}{Fs_{1,1}} \cdot \frac{r^{(2)}s_{2,1}}{Cs_{2,1}}  = \frac{r^{(1)} r^{(2)}}{C F}.
\end{aligned}
\end{equation}

Recall that our goal can be described as the following optimization problem:
\begin{equation}
\begin{aligned}
& \max \rho(r^{(1)},r^{(2)}) \\
&\textrm{s.t.}~~n_{tucker}=C  r^{(2)} + K_1K_2  r^{(1)}  r^{(2)} + F r^{(1)}.\\
\end{aligned}
\label{eq:rank_problem}
\end{equation}

From Eq. \ref{eq:eq3} and the constraint of the above optimization problem, we can have:
\begin{equation}
\begin{aligned}
&\rho = \frac{n_{tucker} - F r^{(1)} - C r^{(2)}}{K_1K_2 C F}\\
\Rightarrow  &\frac{\partial \rho}{\partial r^{(1)}} = \frac{-1}{K_1K_2 C},~~\frac{\partial \rho}{\partial r^{(2)}} = \frac{-1}{K_1K_2 F}.
\end{aligned}
\label{eq:deriv}
\end{equation}

Consider 1) Eq. \ref{eq:deriv} shows that $\rho$ monotonically decreases when $r^{(1)}$ or $r^{(2)}$ increases; and 2) when $r^{(1)}$ or $r^{(2)}$ increases, the constant 
$n_{tucker}$ means the corresponding $r^{(2)}$ or $r^{(1)}$ decreases, increasing the value of $|\Delta Rank|$. We can find that smaller $|\Delta Rank|$ brings larger $\rho$, i.e., more information is preserved after decomposition.  increases. Therefore, when constructing the rank search space, we suggest $|\Delta Rank|$ should be small, e.g., $r^{(1)}=r^{(2)}$.
\begin{algorithm}
\small
\caption{Rank Search Space Determination}
\label{alg:algorithm}
\begin{algorithmic}[1] 
\STATE \textbf{Inputs:} Overall compression ratio $\alpha$, \\
    \quad\quad\quad~ List of the number of input channels $\{C_i\}$, \\
    \quad\quad\quad~ List of the number of out channels $\{F_i\}$,\\
\STATE \textbf{Output:} Low-rank space $\{\{(r^{(1)}_{i,j},r^{(2)}_{i,j})\}\}$.
\\\textcolor[RGB]{0,150,0}{\#~Initialize step size}
\IF {$\max (\{C_i\} \cup \{F_i\}) \ge 128$}
\STATE \textcolor[RGB]{0,150,0}{\#~ Corresponding number of channels is \\ \#~ \{16, 32, 64, 128, 256, 512\}}\\ 
$\{{s_k}\} \gets \{4,8,16,16,32,32\}$
\ELSE
\STATE \textcolor[RGB]{0,150,0}{\#~ For the number of channels less than 128,\\ \#~ the setp size is 4}\\
$\{{s_k}\} \gets \{4\}$ 
\ENDIF\\
\STATE Calculate  $\alpha_{rank}$ by Eq. \ref{eq:r_comp}\\
\textcolor[RGB]{0,150,0}{\#~ Generating rank space for each layer}
\FOR {each $i \in [0,len(\{C_i\})]$}
    \STATE $t \gets min(C_i, F_i)$, ~$k \gets \frac{t}{16}-1$ \\
    \textcolor[RGB]{0,150,0}{\# Relaxing the range of scaling parameter $\alpha_{rank}$}
    \STATE Generate $\{r_j\}$ from [$\frac{t}{\alpha_{rank}+2}$, $\frac{t}{\alpha_{rank}-2}$] with step size $s_k$\\
    \textcolor[RGB]{0,150,0}{\# The size relation between $r^{(1)}$ and $r^{(2)}$ is consistent \\ \# with the original channels, except $r^{(1)} = r^{(2)}$}
    \IF {$C_i \ge F_i$}
    \STATE $m \gets \frac{C_i}{F_i}$
    \STATE $\{(r^{(1)}_{i,j},r^{(2)}_{i,j})\} \gets \{(r_j, \frac{r_j}{m})\} \cup \{(r_j, r_j)\} $
    \ELSE
    \STATE $m \gets \frac{F_i}{C_i}$
    \STATE $\{(r^{(1)}_{i,j},r^{(2)}_{i,j})\} \gets \{(\frac{r_j}{m}, r_j)\} \cup \{(r_j, r_j)\} $
    \ENDIF
\ENDFOR{}
\end{algorithmic}
\end{algorithm}

\end{document}